\documentclass[11pt]{article}
\usepackage{algorithm,fullpage,amsmath,graphicx,amssymb, url}
\usepackage{algpseudocode}
\usepackage{footnote}
\makesavenoteenv{table}
\makesavenoteenv{tabular}
\algnewcommand\algorithmicinput{\textbf{Input:}}
\algnewcommand\algorithmicoutput{\textbf{Output:}}
\algnewcommand\INPUT{\item[\algorithmicinput]}
\algnewcommand\OUTPUT{\item[\algorithmicoutput]}

\begin{document}

\title{Topic Modeling Using Distributed Word Embeddings}

\author{
{\sf Ramandeep S. Randhawa}\thanks{Marshall School of Business, e-mail:
{\tt ramandeep.randhawa@marshall.usc.edu}} \\University of Southern California \and 
{\sf Parag Jain}\thanks{e-mail: {\tt
paragjain78@gmail.com}}\\TopicIQ, Inc.
\and
{\sf Gagan Madan} \thanks{e-mail: {\tt
gagan.madan1@gmail.com}}\\ IIT Delhi}

\date{February 12, 2016}

\maketitle

\begin{abstract}
We propose a new algorithm for topic modeling, \textit{Vec2Topic}, that identifies the main topics in  a corpus using semantic information captured via  high-dimensional distributed word embeddings.  Our technique is unsupervised and generates a list of topics ranked with respect to importance.  We find that it works  better than existing topic modeling techniques such as Latent Dirichlet Allocation for identifying key topics in user-generated content, such as emails, chats, etc., where topics are diffused across the corpus. We also find that \textit{Vec2Topic} works equally well for non-user generated content, such as papers, reports, etc., and for small corpora such as a single-document.
\end{abstract}

%

\section{Introduction}

Understanding an individual's key topics of interest to meet their customized needs is an important challenge for personalization and information filtering applications, such as recommender systems. With the proliferation of diverse types of applications and smart devices, users are generating a large amount of text through their emails, chats, and on social media, such as Twitter and Facebook. This data typically contains many words that reflect their interests: for example, if we consider the work emails of someone who works on datacenter servers, they will likely have several semantically-similar words such as processors, cloud computing, virtualization etc. Individual emails, however, tend to be more about the context in which these words are used, for instance, meetings, status updates, follow-ups, and so on. The topics themselves are diffused across all the emails --- our goal in this paper is to build an unsupervised algorithm that automatically infers these key topics of interest.

State-of-the-art topic modeling algorithms, such as \textit{Latent Dirichlet Allocation} (LDA, \cite{Blei:2003:LDA:944919.944937,Blei:2012:PTM:2133806.2133826}), have been successfully applied to discover the main topics across a large collection of documents for some time now, and are a natural candidate for solving the problem at hand. The typical goal of these methods has been to organize this collection according to the discovered topics --- they work well for news articles, papers in a journal, etc., where the document is about the key topics, and the same words show up several times. User-generated content, however, is usually about the context: actual keywords show up infrequently, and are surrounded by several contextual words such as meetings, status, email, thanks, etc. Because of this difference in structure of corpus, LDA tends to cluster topics with similar contexts, and for each topic tends to capture the most frequently used words. It, however, fails to capture the key topics across the entire corpus.

In this paper, we propose a new technique of topic modeling, \textit{Vec2Topic}, that is aimed toward capturing the key topics across user-generated content. Our algorithm consists of two main ideas. The first idea is that a key topic in a corpus should contain a large number of semantically-similar words that relate to that topic. We capture this notion using a \textit{depth measure} of a word, that helps identify clusters with the highest density of semantically-similar words. We leverage word semantics as captured by high-dimensional distributed word embeddings. Specifically, we perform agglomerative clustering on these distributed word vectors over the vocabulary, and we use the resulting dendrogram to compute the depth score for a word as the number of links between the word and the root node. Clusters that contain words with the highest depth scores reflect the user's key topics of interest.   Our second idea is aimed at deriving good labels that best describe these key topics --- the idea is that such keywords not only have high depth, but also show up in the context of a large number of different words. We capture this using a \textit{degree measure} of a word, which we define as the count of the number of unique words that co-occur with the word within a context window in the corpus. We then combine the depth and degree measures into a single \text{word score}. We use this score to rank all the topics in the corpus.
\begin{table}
\centering
\caption{Former Enron CEO Kenneth Lay's emails: Top-10 words from the Top-3 scoring topics in \textit{Vec2Topic} compared with three LDA-based topics}
\begin{tabular}{c c c || c c c} \hline
\multicolumn{3}{c||}{\textit{Vec2Topic}}&\multicolumn{3}{c}{\textit{LDA}}\\
Topic 1 & Topic 2 & Topic 3& Topic A & Topic B& Topic C\\ \hline
business   	&	power		&	strategy	&meeting &	senate &	conflict\\
development   	&	supply		&	product		& email &	tax &	libya\\
market 		&	cost		&	internet	& time &	sept &	developer\\
management   	&	risk		&	solutions	& information &	gift &	end\\
gas   		&	demand		&	service & business &	speaker &	context\\
oil   		&	california	&	tax		& company &	truth &	delegate\\
technology   	&	electricity	&	company		& thanks &	computer &	conf\\
policy 		&	emission	&	software	& conference &	arrival &	guide\\
industry   	&	reduction	&	marketing	& market &	congressman &	gap\\
future   	&	system		&	budget & year &	result &	click\\
\hline
\end{tabular}
\label{tab:lay}
\end{table}

\begin{table}
\centering
\caption{Former Enron Managing Director of Research V. Kaminski's emails: Top-10 words from the Top-3 scoring topics in \textit{Vec2Topic} compared with three LDA-based topics}
\begin{tabular}{c c c || c c c} \hline
\multicolumn{3}{c||}{\textit{Vec2Topic}}&\multicolumn{3}{c}{\textit{LDA}}\\
Topic 1 & Topic 2 & Topic 3 & Topic A & Topic B & Topic C\\ \hline
analysis&	investment&	government&	enron&	conjecture&	interest\\
model&	market&	commission&	thanks&	idg&	increment\\
approach&	equity&	debt&	email&	eia&	optimisation\\
probability&	asset&	tax&	market&	involvement&	garp\\
modeling&	insurance&	authority&	time&	arkansas&	klaus\\
calculation&	banking&	policy&	price&	luncheon&	freebie\\
method&	bank&	payment&	company&	piano&	plateau\\
estimation&	capital&	requirement&	need&	fire&	uhc\\
volatility&	markets&	regulation&	work&	protection&	virtue\\
pricing&	technology&	ferc&	year&	answer&	regression\\
\hline
\end{tabular}
\label{tab:kaminski}
\end{table}

To illustrate this concept, we ran \textit{Vec2Topic} on Enron's publicly available email corpus. Table~\ref{tab:lay} shows the results on former Enron CEO Kenneth Lay's  emails.  Topics 1-3 are the top-3 topics identified by \textit{Vec2Topic} in decreasing order of importance; each topic is listed with the ten highest scoring constituent words. We can see that the most important topic is about the corporate functions of a CEO (business, development, market, policy, etc.), the next topic reflects that Enron is an energy company (power, supply, demand, electricity, etc.), and the third topic is more about its operating aspects (product, strategy, marketing, budget, etc.). We then contrast it with LDA\footnote{We ran LDA with $K=10,25,$ and $50$ and picked the topics that seemed most relevant.}, which tends to cluster emails with similar context into topics. In Table~\ref{tab:lay}, we notice that LDA's Topic $A$ captures words like meeting, email, time, thanks, etc., that show up frequently in the context of his emails; Topic $C$ seems to capture a specific theme on the Libya conflict. While these are themes in Mr.  Lay's emails, they do not capture the overall essence of who Mr. Lay is, what he works on, and what his key interests are across all his communication. \textit{Vec2Topic} does a much better job in capturing this insight, and in a sense is answering a different question than LDA does.  Table~\ref{tab:kaminski} shows similar results on Enron's Former Managing Director of Research, V. Kaminski's emails --- again, we see that it captures key topics such as modeling and analysis, investment, and government-related topics, which are fairly reflective of his work portfolio.

To summarize, we propose a new architecture for topic modeling that is geared for extracting key topics of interest from user generated text content. We show that it works better than existing topic modeling techniques such as LDA.  We also study its performance on non-user generated content, and find that it works equally well on such datasets. In particular, we show results on the set of all the accepted papers from NIPS 2015 conference. We also demonstrate its efficacy when run on a single-document by  running it on the annual financial report of Apple.

\section{Background}
\label{sec:background}
In 2003,  LDA was introduced as a generative probabilistic topic modeling model to manage large document archives, \cite{Blei:2003:LDA:944919.944937}. The goal was to discover the main themes across a large collection of documents, and organize the collection according to the discovered themes. The intuition behind LDA is that documents have a latent topic structure; each document is  a probability distribution over topics, and each topic in turn is a distribution over a fixed vocabulary. Words in the vocabulary are represented by 1-of-$V$ encoding, where $V$ is the size of the vocabulary --- there is no notion of semantic similarity between words. The central problem is to use the observed documents to infer the hidden topic structure.

Since then, a huge body of work has been done to relax and extend the statistical assumptions made in LDA to uncover more sophisticated structure in the text: Topic models that assume that topics generate words conditional on the previous word \cite{wallach2006topic}; dynamic topic models that respect the ordering of documents \cite{blei2006dynamic}; Bayesian non-parametric topic models that generate a hierarchy of topics from the data  \cite{blei2010nested, teh2006hierarchical}; Correlated topic models that allow topics to exhibit correlation \cite{blei2007correlated}; Relational topic models that capture both the topic model and a network model for the documents  \cite{chang2010hierarchical}.  

Our algorithm is inspired by the recent work in learning word vector representations using neural networks \cite{Bengio:2003:NPL:944919.944966,Collobert:2008:UAN:1390156.1390177,mnih2009scalable,turian2010word,Collobert:2011:NLP:1953048.2078186}. In this formulation, each word is represented by a vector that is concatenated or averaged with other word vectors in a context, and the resulting vector is used to predict other words in the context. The outcome is that after the model is trained, the word vectors are mapped into a vector space such that semantically-similar words have similar word representations (e.g. ``strong'' is close to ``powerful''). \cite{Bengio:2003:NPL:944919.944966} used a feedforward neural network with a linear projection layer and a non-linear hidden layer to jointly learn the word vector representations and a statistical language model. \cite{Collobert:2008:UAN:1390156.1390177} and \cite{Collobert:2011:NLP:1953048.2078186} leveraged distributed word vectors to show that neural network based models match or outperform feature-engineered systems for standard Natural Language Processing (NLP) tasks  that include part-of-speech tagging, chunking, named entity recognition, and semantic role labeling.  \cite{Huang:2012:IWR:2390524.2390645} introduced a technique to learn better word embeddings by incorporating both local and global document context, and account for homonymy and polysemy by learning multiple embeddings per word.

\cite{DBLP:journals/corr/abs-1301-3781,NIPS2013_5021,mikolov2013linguistic} introduced \textit{Word2Vec} and the \textit{Skip-gram} model, a very simple method for learning word vectors from large amounts of unstructured text data. The model avoids non-linear transformations and therefore makes training extremely efficient. This enables learning of high-dimensional word vectors from huge datasets with billions of words, and millions of words in the vocabulary. High-dimensional word vectors can capture subtle semantic relationships between words. For example, word vectors can be used to answer analogy questions using simple vector algebra: $v_{King} -v_{man} + v_{woman} = v_{Queen}$, where $v_x$ denotes the vector for the word $x$. \cite{pennington2014glove} later introduced \textit{GloVe}, Global Vectors for Word Representation, which combines word-word global co-occurrence statistics from a corpus, and context based learning similar to Word2Vec to deliver an improved word vector representation. Word vectors are an attractive building block and are being used as input for many neural net based natural language tasks such as sentiment analysis \cite{socher2011dynamic, socher2011parsing,socher2011semi,socher2013recursive}, question and answer systems, \cite{kumar2015ask,weston2015towards}, and others.

\section{Methodology}
In this section, we describe our algorithm for identifying the key topics underlying a corpus.  Our approach is three-fold.  First, we build distributed word embeddings for the vocabulary  of the corpus. Second, we cluster the word embeddings using $K$-means to yield $K$ clusters of semantically-related words.  We implement $K$-means using the standard euclidean distance metric, but with all word vectors  normalized so that their norm is unity. We identify each of the $K$ clusters obtained through $K$-means as a topic.  Third, we score the importance of each topic and label the keywords that best describe the topic. 

The core of \textit{Vec2Topic} lies in the third step, i.e., scoring the importance of topics.  So, we will focus on describing this next in Section~\ref{sec:metric}. For ease of exposition, we proceed by assuming that the first and second steps have been completed. That is,  we are in possession of good distributed word embeddings for all nouns and nouns phrases in the corpus, and that these have been clustered using $K$-means. We formally discuss how to build the word embeddings in Section~\ref{sec:embeddings}. Algorithm~\ref{topic_algo} describes all steps of \textit{Vec2Topic}.
\\

\subsection{Scoring importance of topics}
\label{sec:metric}
We put forth the basic idea that: \textit{ a core topic in a corpus should contain a large number of words that relate to that topic, and further, these words, though distinct, should have similar meanings.} For instance, we expect the email corpus of Mr. Kaminski (who was Enron's Managing Director of Research) to contain many modeling-related words such as ``probability'', ``standard deviation,'' ``covariance'', etc. Further, when considered in the vector space of the word embeddings, these words should form a tight cluster with a large number of words that are near each other. We do expect additional clusters to form, for instance, time-related words such as days of the week and names of months would have vector representations that would be close to each other. However, such a time-related cluster would be much less dense than the modeling-related one because modeling would consist of a much larger number of words that are closer in the high-dimensional space. In this fashion, identifying the densest clusters can help us identify the core topics.

To formally capture this notion of cluster density, we use the notion of \textit{depth} of a word. In particular, we take all the words in our vocabulary and perform hierarchical (agglomerative) clustering on the corresponding word vectors. This is an iterative method that starts by placing all words in their own cluster, and then the proceeds by merging the two closest clusters in each step, until a single cluster that includes all words is obtained. We use the typical \text{cosine}  distance measure for this clustering approach, i.e., for any two word vectors $w_x$ and $w_y$, we define the distance
\begin{equation}
d(w_x,w_y)=1-\frac{w_x}{\parallel w_x \parallel_2 } \cdot \frac{w_y}{\parallel w_y\parallel_2 },
\end{equation}
where $\parallel\cdot\parallel_2$ denotes the Euclidean norm. As an output of this clustering method, we obtain a dendrogram, which is a tree with each word represented by a leaf node and every non-leaf node representing a cluster that contains the words corresponding to its leaf nodes. The root node of the dendrogram reflects the cluster of \textit{all} words. We modify this dendrogram by normalizing the length of all links between parent and children nodes to unity. We then define the depth of a word as the (minimum) distance of the (leaf) node that corresponds to the word from the root node on this modified tree. Thus, the depth of a word can also be understood as the minimum number of links that need to be traversed to travel from the word node to the root node. That is,
\begin{equation}
\label{eq:depth}
depth(w)=\text{No. of links between root node and $w$}.
\end{equation}
Table~\ref{tab:depth} lists the ten deepest words for Mr. Kaminski's email corpora.

\begin{table}
\centering
\caption{Mr. Kaminiski's email corpus: Top-10 words based on Depth and Score}
\begin{tabular}{c c} \hline
Depth & Score\\
\hline
approximation&	analysis\\
probability&	model\\
covariance&	approach\\
estimation&	probability\\
variance&	modeling\\
calculation&	calculation\\
convolution&	investment\\
parameter estimation&	method\\
standard deviation&	market\\
coefficient&	estimation\\
\hline
\end{tabular}
\label{tab:depth}
\label{tab:metric}
\end{table}

\begin{figure}
\begin{minipage}[b]{0.45\linewidth}
\centering
\includegraphics[scale=0.2]{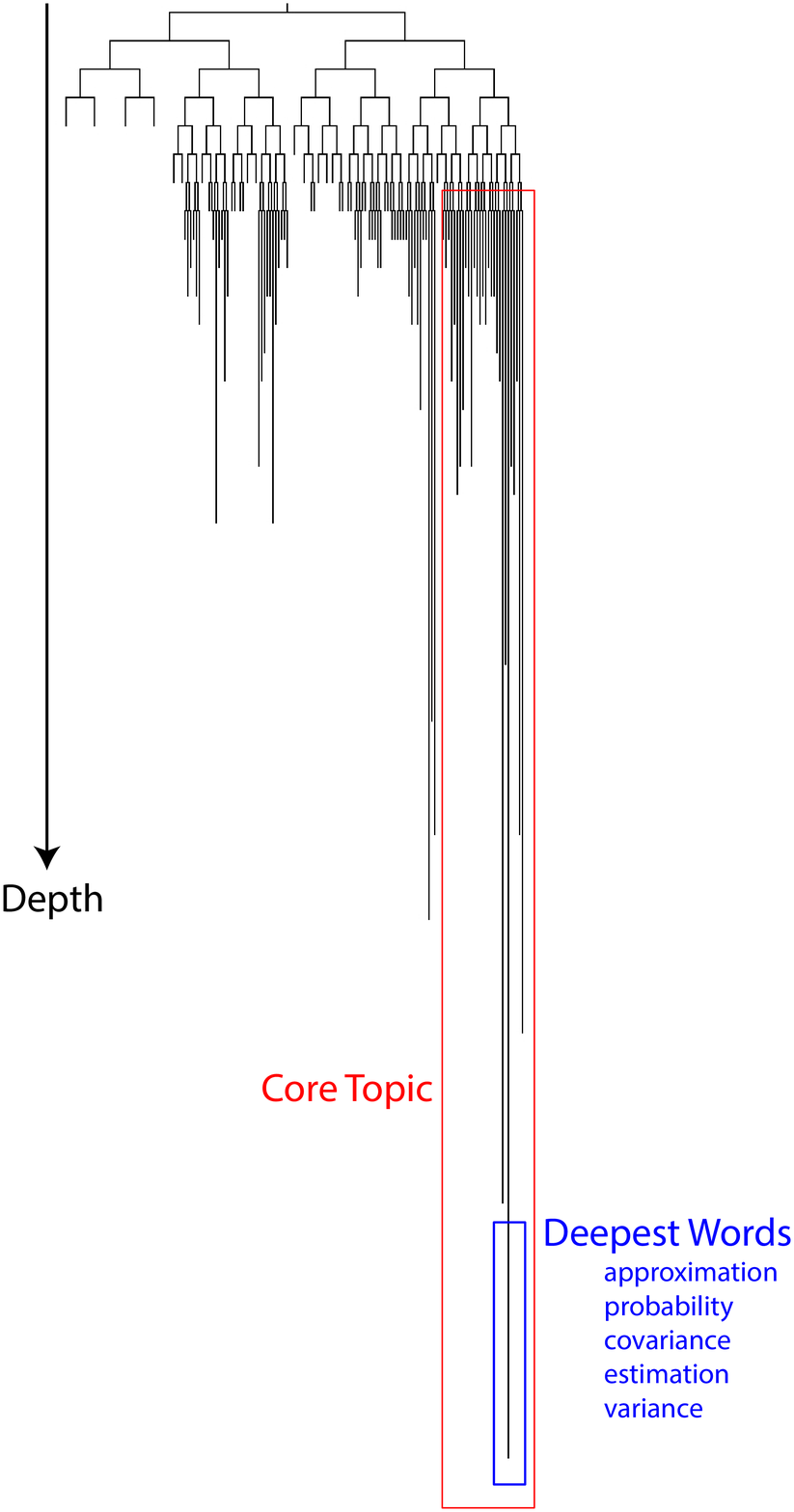}
\label{fig:dendrogram}
\end{minipage}
\quad
\begin{minipage}[b]{0.45\linewidth}
\centering
\includegraphics[scale=0.3]{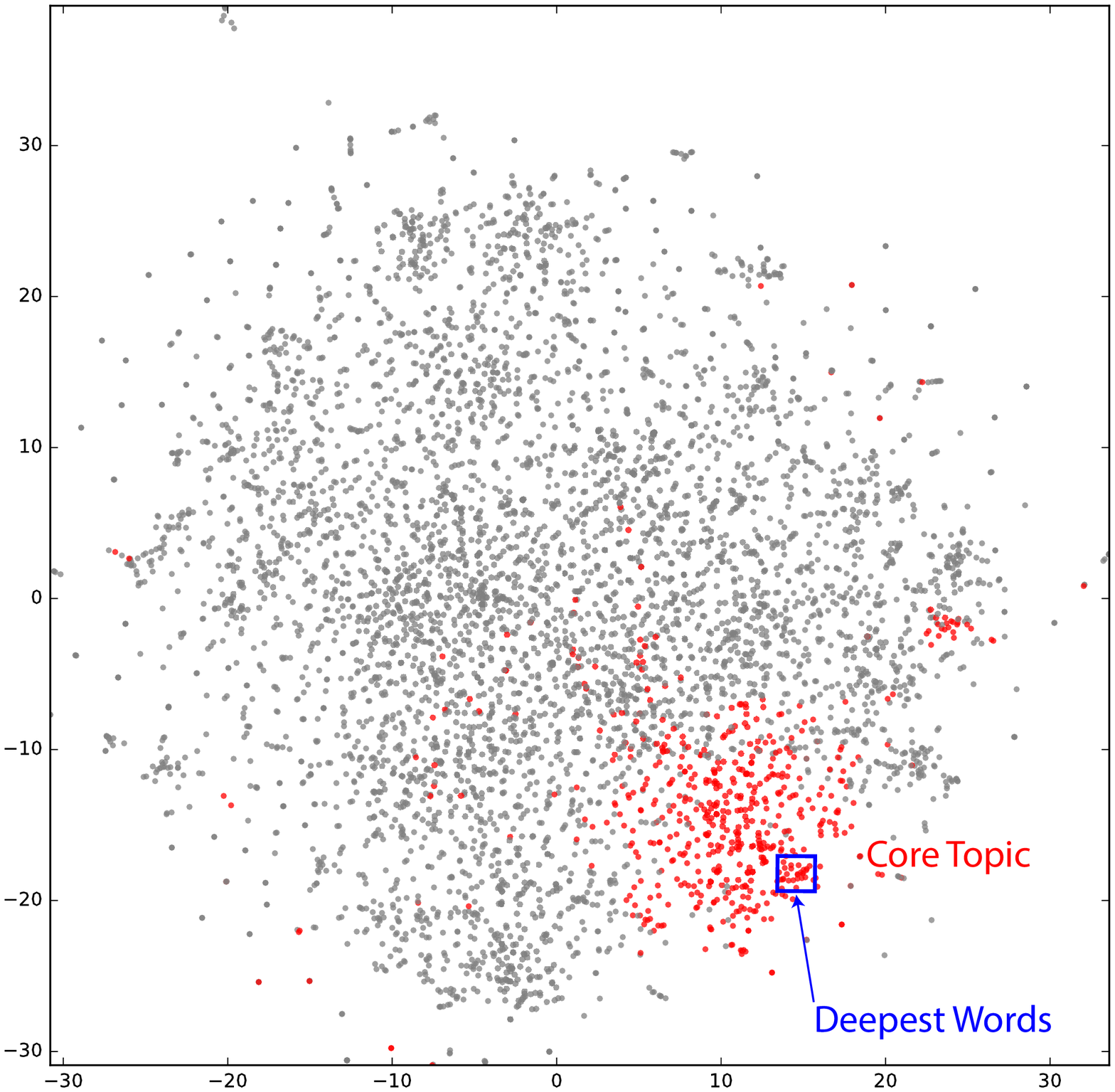}
\label{fig:tsne_depth} 
\end{minipage}
\caption{Mr. Kaminski's vocabulary: (Left) Depth view of hierarchical clustering and (Right) Word vectors plotted using t-SNE. The blue box encloses the deepest words, and the red box (Left) and red colored points (Right) indicate the core topic that contains the deepest words.}
\label{fig:tree}
\end{figure}

We next explain this measure in more detail by focusing on Mr. Kaminski's email corpus.  Figure~\ref{fig:tree} (left) displays the modified  dendrogram corresponding to the agglomerative clustering over the vocabulary. This dendrogram preserves the clusters of the agglomerative clustering approach, but creates links with unit length at each level. At the top of the dendrogram is the root node; words that cluster close to the root do not have many semantically-similar words close to them in vector space, and hence have low depth. The words with the highest depth are the ones that have many semantically-similar words that are close to them, and therefore, are farthest away from the root. The topic cluster that the deepest words belong to reflects the core topic for the corpus. 

Figure~\ref{fig:tree} (right) displays our word embeddings in two-dimensional space. We clearly see that words with the highest depth are very close to each other in vector space, and the topic they belong to is the core topic. Our word embeddings are 325-dimensional vectors (which will be described in more detail in Section~\ref{sec:embeddings}) so visualizing them in two-dimensions perfectly is not possible. However, we use the t-SNE technique \cite{tsne:08}, which does a remarkable job of dimension-reduction by trying to ensure that the distances in two dimensions are reflective of those in the higher-dimensional space. 

We notice that the depth measure helps identify the core topics, however, the deepest words are by themselves not sufficient to understand or label the overall topic to which they pertain. For instance,  Mr. Kaminski's deepest words are ``approximation, probability, covariance,'' which are quite specialized and do not explicitly reveal their overall topic.

To identify words that provide good labels for the  key topics, we next put forth our second idea: \textit{topical words that are indicative of core topics would be used in the context of a large number of different words}. For instance, a modeling-related topic in Mr. Kaminski's email corpus should contain ``model'' or a similar word as a topical word, and this word should be used in the context of many words, which would indicate that Mr. Kaminski is communicating about different types of models, with different people, etc. We use the degree of a word as another independent measure that quantifies this notion of ``topicality.'' We compute the degree of the words as follows: we build a graph of the corpus by dividing the corpus into individual sentences; we create a link between any two words of the vocabulary $V$ that co-occur in the same sentence; then, we define  the number of neighbors of each word in this graph as the degree of that word. That is,
\begin{equation}
\label{eq:degree}
degree(w)=\text{No. of unique words co-occurring with $w$}.
\end{equation}

We next combine the depth and degree measures in \eqref{eq:depth} and \eqref{eq:degree} to create an overall score for each word. We do so by normalizing depth and degree measures by their maximum values across the vocabulary. The degree measure tends to be quite skewed, so we take a logarithmic transformation before scaling by the maximum. Formally, the score is defined as: for $v\in V$,
\begin{equation}
\label{eq:score}
\begin{split}
Score(v)=&\left(\frac{depth(v)}{\max_{u\in V}depth(u)}\right)^\alpha  \\ 
\times&\left(\frac{\log(1+degree(v))}{\log(1+\max_{u\in V}degree(u))}\right)^\beta,
\end{split}
\end{equation}
where $\alpha$ and $\beta$ are normalization parameters; in all our experiments we set $\alpha=\beta=1$ and we discuss a generalization of this in Section~\ref{sec:normalization}. Intuitively, the deepest words pertain to core topics but can be too specialized to indicate the topic. So, to identify the topics, we need to identify words that are similar in meaning to the deep words but are also used in various contexts. That is, these words should have both high depth and high degree. The score in \eqref{eq:score} reflects this intuition by formally multiplying the two measures, after scaling them appropriately. Note that because we take the logarithmic transformation of degree, we consider $(1+degree(v))$ to avoid the argument of the logarithm from taking the value zero.  

Table~\ref{tab:metric} displays the top 10 words based on this score for Mr. Kaminski's corpus. We believe that this list of words seems quite consistent with what one would expect to be his important words and provides labels for the core topics appropriately. For instance, ``analysis'' and ``model'' are the top words, which represent the core topic of modeling quite well.

\begin{figure}
\centering
\includegraphics[scale=0.3]{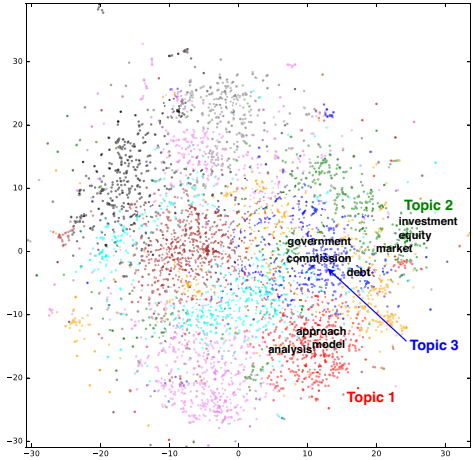}
\caption{Visualizing the topics captured from Mr. Kaminski's corpus using t-SNE; Top-3 words of the top-3 topics are marked.}
\label{fig:tsne}
\end{figure}

Once the words have been scored, we compute a score for each topic by averaging the score of the words that comprise the corresponding cluster (based on $K$-means clustering). That is,
\begin{equation}
\label{eq:topic_score}
  \text{Score of }Topic_i=\frac{\sum_{v \in Topic_i} Score(v)}{|Topic_i|}.
\end{equation}
Thus, we obtain a sorted list of topics, with higher scoring topics identified as being more important than lower scoring ones. Tables~\ref{tab:lay} and \ref{tab:kaminski} depict the topics with the highest three scores from Mr. Lay's and Mr. Kaminski's corpora, respectively. In both cases, the
topics are numbered in decreasing order of the score; Topic 1 has the highest score. Notice that within each topic, the constituent words are also ranked based on the word score. In this manner, each topic is represented by the words most relevant to that topic. Figure~\ref{fig:tsne} visualizes $K = 10$ topics using the vocabulary of Mr. Kaminski's emails. The ten clusters are labeled in different colors; the top-3 topics are numbered as in Table~\ref{tab:kaminski}  and their top-3 words are also displayed.

\begin{algorithm*}
\caption{\textit{Vec2Topic}}
\label{topic_algo}
\begin{algorithmic}[1]
\INPUT{}
\Statex Text corpus
\Statex Knowledge-base $k$ of word embeddings
\Statex Number of topics to extract, $K$
\item[{\textbf{Build word embeddings:}}]
\State Extract vocabulary of nouns and noun phrases $V$ from corpus
\State Learn distributed word representations on corpus, $\ell_v$ for $v\in V$
\State Compute word vectors $w_v=[k_v;\ell_v]$ for $v \in V$
\item[{\textbf{Identify topics:}}]
\State Perform $K$-Means clustering to obtain $Topic_i$ for $i=1,\dots,K$
\item[{\textbf{Score topics:}}]
\State Perform agglomerative clustering on $\{w_v:v \in V \}$
\State Compute $depth(v)$ for $v\in V$ using \eqref{eq:depth}
\State Compute the degree of each word, $degree(v)$ for $v \in V$, using the sentence-level co-occurrence graph as in \eqref{eq:degree}.
\State For $v\in V$, compute:
$$Score(v)=\left(\frac{depth(v)}{\max_{u\in V}depth(u)}\right)^\alpha \left(\frac{\log(1+degree(v))}{\log(1+\max_{u\in V}degree(u))}\right)^\beta,$$
where $\alpha$, $\beta$ are normalization parameters; all our experiments use $\alpha=\beta=1$.
\State Compute average score of each topic as the average score of constituent words using \eqref{eq:topic_score}
\OUTPUT{} 
\Statex $Score(v)$ for $v \in V$: the score for each word
\Statex $Topic_{[i]}$ for $i=1,\dots,K$: the ranked list of topics
\end{algorithmic}
\end{algorithm*}

\subsection{Building distributed word embeddings}
\label{sec:embeddings}

A natural way to build distributed word embeddings is to apply a standard technique such as skip-gram \cite{DBLP:journals/corr/abs-1301-3781} on the given corpus. This approach however assumes that the training corpus  is large enough, and words show up in several different contexts to develop semantically accurate word representations; trained models have been built on corpuses that have billions of tokens and millions of words in the vocabulary \cite{DBLP:journals/corr/abs-1301-3781,NIPS2013_5021}. User generated data, however, tends to have much smaller corpus and vocabulary size, which does not allow understanding the relative meaning of words sufficiently; nevertheless, training this model on the data does capture how these words are used in the user's context. 

To overcome this limitation, we take a two step approach: we learn ``global'' word embeddings using the skip-gram technique on a knowledge-base  --- these word embeddings tend to capture the generic meaning of words in widely used contexts. We then augment these with ``local'' word embeddings learned from the user's data to capture their context to generate our desired word vectors. Denoting the knowledge-base vectors by $k:\mathcal{V}\rightarrow \mathbb{R}^{\kappa}$, where $\mathcal{V}$ is the knowledge-base vocabulary,  and our local  word vectors by $\ell: V \rightarrow \mathbb{R}^{\lambda}$, we use the concatenated word vectors $w=[k(v);\ell(v)]$ for $v\in V$, with $w(v)\in \mathbb{R}^d$, and $d:=\kappa+\lambda$. For our experiments, we first use the skip-gram technique on the Wikipedia corpus that contains about three million unique words, and about six billion total tokens. We set $\kappa=300$ and $\lambda=25$ to obtain 325-dimensional word vectors after combining the knowledge-base and local word vectors. 

It is educational to consider the cases in which the augmentation is not performed and the word vectors are derived either solely based on the given corpus, or solely based on the knowledge-base. Table~\ref{tab:concat} presents the top four topics obtained for both of these cases on Mr. Kaminski's email corpus. Comparing the two cases, we see that when using the local word vectors alone, word semantics are not captured that well, and thus the topics obtained are more diffuse. When using knowledge-base word vectors, the semantics dominate the topic definition with similar meaning words pulled into topics from all over the corpus, without consideration for the context. When one augments both these word vectors, we obtain a good balance that yields the sharper results of Table~\ref{tab:kaminski}.

\begin{table*}
\centering
\caption{Mr. Kaminski's email: Comparing results of local word vectors with knowledge-base word vectors}
\resizebox{\columnwidth}{!}{
\begin{tabular}{c c c c ||c c c c  } \hline
\multicolumn{4}{c||}{Local word vectors}& \multicolumn{4}{c}{Knowledge-base word vectors}\\ 
Topic 1 & Topic 2 & Topic 3 & Topic 4& Topic 1 & Topic 2 & Topic 3 & Topic 4\\
\hline
state &	model &	microsoft &	bank &	investment &	microsoft &	analysis &	morgan\\
government &	volatility &	package &	banking &	volatility &	server &	process &	harvey\\
dpc &	market &	user &	officer &	equity &	user &	rate &	campbell\\
dabhol &	option &	vulnerability &	equity &	pricing &	desktop &	method &	richardson\\
demand &	approach &	server &	investment &	valuation &	email &	calculation &	henry\\
commission &	pricing &	machine &	division &	company &	interface &	modeling &	collins\\
mseb &	spot &	code &	bond &	asset &	unix &	function &	morris\\
maharashtra &	var &	source &	morgan &	payment &	data &	estimation &	freeman\\
authority &	curve &	world &	brokerage &	transaction &	browser &	methodology &	fisher\\
supply &	valuation &	flaw &	march &	investor &	freebsd &	curve &	massey\\
\hline
\end{tabular}
}
\label{tab:concat}
\end{table*}

\section{Discussion}
In this section, we discuss various aspects of  \textit{Vec2Topic}. First, we discuss its speed in Section~\ref{sec:complexity}, then, in Section~\ref{sec:K}, we discuss its robustness to  the number of topics to extract. In Section~\ref{sec:size} we demonstrate its performance  when implemented on: a non-user generated dataset (NIPS 2015 papers) and a small dataset (a single-document, Apple's 10-K financial report) . Finally, in Section~\ref{sec:implementation}, we discuss some additional considerations in implementing the algorithm.

\subsection{Algorithm complexity}
\label{sec:complexity}
To understand the complexity of the algorithm, we consider each of its key components. We use $\bar{V}$ to denote the entire vocabulary of the corpus (recall that $|V|$ is the vocabulary of nouns and noun phrases).
\begin{enumerate}
\item \textbf{Hierarchical (agglomerative) clustering}: we use the \text{fastcluster} method, \cite{mullner2013fastcluster}, which has a complexity of $\Theta(|V|^2)$. 
\item \textbf{Building word vectors}: we use the skip-gram model \cite{DBLP:journals/corr/abs-1301-3781}, which has a running complexity of $E\times T \times Q$,  where $E$ is the number of iterations, which is typically $5-50$, $T$ is the total number of words or tokens in the corpus, and $Q=c \times (x+x \log_2(|\bar{V}|))$, where $c$ is the context size (in all our experiments we set $c=5$) and $x\in\{\kappa,\lambda\}$ denotes the dimensionality of the word vectors used in training. The case $x=\kappa$ represents the training over the knowledge-base, which needs to only be done once or a limited number of times, and $x=\lambda$ represents the training over the local corpus.
\item \textbf{$K$-means clustering}:  this has a complexity of $O(|V| \times K \times d\times i)$, where $i$ is the number of iterations.
\item \textbf{Computing Score}: The score has two components: depth and degree. Computing depth is linear in the size of the vocabulary of interest  $|V|$ once the hierarchical clustering has been done. Computing degree involves building a co-occurrence matrix for the entire vocabulary $\bar{V}$ over each sentence, and this has a worst case complexity of $O(|\bar{V}|^2 \times S)$, where $S$ is the  number of sentences in the corpus.
\end{enumerate}

\begin{table}
\centering
\caption{Details of all corpora studied}
\begin{tabular}{c c c  c c  } \hline
Corpus & Documents& Tokens & $|V|$ & Runtime\footnote{Based on a 4-core 4GHz Intel processor with 32GB RAM. Python 2.7 was used on a Ubuntu machine with the libraries Numpy, Scipy, Scikit-learn (for K-means clustering), Fastcluster (for agglomerative clustering), and Gensim (for extracting bigrams and running Skip-gram); many components were not parallelized. The runtime is the total time after text corpus is presented as input and includes time for any pre-processing. Our pre-processing included running a lemmatizer on the corpus, and in the case of an email corpus,  removing any names that were included in \textit{to, from, cc} fields.  Note: runtime excludes the time taken to load the knowledge-base vectors into memory, which was about 40 seconds.} \\
\hline
Apple 10-K&1&58,795&632& $\sim$4 sec\\
Lay & 540 & 104,194 &  1,301& $\sim$7 sec\\
Kaminski & 8,644 & 1,171,632& 5,739&  $\sim$70 sec\\
NIPS 2015 &403&1,943,649& 4,754&$\sim$ 100 sec\\
\hline
\hline
\end{tabular}
\label{tab:corpus_details}
\end{table}

In all our experiments, the algorithm does not take more than a few minutes to run. 
Table~\ref{tab:corpus_details} provides details on the sizes of all the datasets considered in this paper and the runtime of the algorithm. 
The slowest component of  \textit{Vec2Topic} is  building word vectors, which took  majority of the run time. However, this method is quite scalable and has been used to build word vectors on a Google news dataset of 100 billion tokens in a couple of days. Indeed, we trained our knowledge-base vectors on Wikipedia corpus, which contains about 6 billion tokens and about 2.8 million unique words in about 6 hours. 

The $K$-means clustering has an efficient implementation as well, and scales with sizes well. Further, computing the degree measure can also be done very efficiently as it involves hashing as well as sparse matrix operations. So, as the scale of the corpus grows, we believe the critical component of the algorithm from a run time perspective is hierarchical clustering. For $|V|=5,000$ with $d=325$, this component completes in about 2.5 seconds. This scales up quadratically so that $|V|=10,000$ takes 10 seconds, and $|V|=50,000$ takes slightly more than four hours. Given that $V$ only comprises nouns and noun phrases, we believe that its size will be an order smaller than the total unique words in the document, and thus overall, the algorithm has the ability to scale well with data size.

\begin{table*}
\centering
\caption{Mr. Kaminski's email: Top-4 topics for different $K$ values}
\resizebox{\columnwidth}{!}{
\begin{tabular}{c c c c ||c c c c  } \hline
\multicolumn{4}{c||}{$K=5$}& \multicolumn{4}{c}{$K=50$}\\ 
Topic 1 & Topic 2 & Topic 3 & Topic 4& Topic 1 & Topic 2 & Topic 3 & Topic 4\\
\hline
analysis &	investment &	server &	thing &		analysis &	investment &	market &	process\\
model &	market &	application &	something &		model &	equity &	technology &	testing\\
approach &	equity &	microsoft &	everyone &		approach &	asset &	development &	implementation\\
probability &	demand &	technology &	anything &		probability &	insurance &	customer &	assessment\\
modeling &	asset &	desktop &	someone &		modeling &	banking &	industry &	evaluation\\
calculation &	growth &	internet &	anyone &		calculation &	bank &	internet &	concept\\
method &	prices &	enterprise &	fact &		method &	capital &	enterprise &	requirement\\
estimation &	insurance &	browser &	lot &		estimation &	markets &	ecommerce &	framework\\
volatility &	banking &	user &	everything &		volatility &	company &	companies &	reliability\\
pricing &	bank &	ecommerce &	nothing &		pricing &	investor &	solution &	monitoring\\
\hline
\end{tabular}
}
\label{tab:K}
\end{table*}

\subsection{Effect of changing $K$}
\label{sec:K}
Similar to other topic modeling methods, our algorithm takes as input the number of topics to be extracted, $K$. For the results described so far, we  fixed $K=10$. Table~\ref{tab:K} displays the top-four topics for Mr. Kaminski when the number of topics $K$ is set to 5 and 50. Notice that in both cases, the overall word score remains unaffected by the change in $K$, the only change is the clustering of topics. Observe that the top topic is identical for these two cases, and for the case $K=10$ of Table~\ref{tab:kaminski}. Even the second-ranked topic is quite similar. This illustrates the robustness of our approach in capturing the key topics from a corpus.

Notice that the case $K=5$ is quite crude in the sense that all words are clubbed into 5 clusters and hence we observe a diffuse topic such as Topic 4. On the other hand $K=50$ is much more pointed and picks up a large number of small clusters, which leads to identifying many specialized topics. In our extensive experiments, we found that moderate $K$ values, such as  $K=10$ or 20, work well in identifying key topics.

\subsection{Other datasets}
\label{sec:size}
\paragraph{NIPS 2015 Dataset}
We next consider a dataset consisting of the full-text of all papers accepted at the NIPS 2015 conference.\footnote{We used Andrej Karpathy's scripts available at \url{https://github.com/karpathy/nipspreview.git} to scrape this information. As part of pre-processing we removed the References section of each paper; we discuss the value of doing so in Section~\ref{sec:implementation}.} Table~\ref{tab:nips} lists the top-3 topics obtain from applying  \textit{Vec2Topic} compared with three relevant LDA-based topics (we ran LDA with $K=10$ to be consistent with our approach). We note that in this case LDA does a very good job of capturing topics. In fact, the table provides a good insight into how  \textit{Vec2Topic} differs from LDA. Our algorithm focuses on concepts \textit{across} the documents in the dataset, whereas LDA appears to focus on concepts \textit{within} each document in the dataset. Consider Topic 1 and Topic $A$, which contain two common words (``graph'' and ``matrix''). Topic 1 from  \textit{Vec2Topic} places these words along with others that tend to be used in similar themes, such as graph theory. Although Topic $A$ also appears to relate to the topic of graph theory, it is more focused on the context of each paper, and contains broader words such as ``data'' and ``problem.'' Again Topic 2 and Topic $B$ appear to be about inference, but Topic 2 contains terms that relate to inference such as ``algorithm'', ``MCMC'', whereas Topic $B$ seems to be more about the context in which inference is done in the papers with words: ``submodular'', ``function'', ``greedy'', etc. Finally, Topic 3 is about general concepts such as ``problem'',  ``complexity'', ``efficiency'', etc., that is once again a common topic across all the papers, whereas Topic $C$ appears to be about learning algorithms.

\begin{table}
\centering
\caption{NIPS 2015 Dataset: Top-3 scoring topics in  \textit{Vec2Topic} compared with three LDA-based topics}
\begin{tabular}{c c c || c c c} \hline
\multicolumn{3}{c||}{\textit{Vec2Topic}}&\multicolumn{3}{c}{\textit{LDA}}\\
Topic 1 & Topic 2 & Topic 3 & Topic A & Topic B & Topic C\\ \hline
graph&	algorithm&	problem&	algorithm&	set&	algorithm\\
approximation&	bayesian&	idea&	sparse&	algorithm&	regret\\
gaussian&	inference&	complexity&	set&	submodular&	bound\\
subspace&	regression&	notion&	graph&	problem&	loss\\
nonzero&	optimization&	efficiency&	matrix&	function&	learning\\
eigenvalue&	computation&	interpretation&	data&	time&	online\\
equation&	classifier&	proof&	random&	number&	bounds\\
manifold&	mcmc&	knowledge&	problem&	inference&	problem\\
covariance&	bayesian inference&	concept&	theorem&	greedy&	function\\
matrix&	speedup&	sense&	number&	node&	convex\\
\hline
\end{tabular}
\label{tab:nips}
\end{table}

\paragraph{Apple's 2015 10-K Financial Report} We next apply  \textit{Vec2Topic} to a single document. We consider Apple's financial report, the 10-K document for 2015. We apply  \textit{Vec2Topic} with $K=10$. Table~\ref{tab:apple} displays the top-4 topics from the report along with the top-10 highest scoring words. We notice that Topic 1 corresponds to the financial accounting terminology that one expects to be central to the financial report (asset, tax, income, etc.). Topic 2 is about the nature of the company (produce, software, hardware, etc.); Topic 4 provides specifics on these (iOS, iPhone, iPad, etc.). Topic 3 pertains to financial considerations of the company (cost, risk, volatility, etc.). Notice that the list of top words goes across the different topics and contains both financial and product-related words (asset, product, software, income, etc.). We would like to point out that LDA does not directly apply to a single document so we do not provide the comparison.

\begin{table}[h]
\centering
\caption{Apple 2015 10-K: Top-10 scoring overall words and Top-4 scoring topics in \textit{Vec2Topic}} 
\begin{tabular}{c ||c c c c  } \hline
Top words&	Topic 1&	Topic 2&	Topic 3&	Topic 4\\ \hline
asset&	asset&	product&	cost&	ios\\
product&	tax&	software&	risk&	iphone\\
tax&	income&	hardware&	expense&	ipad\\
software&	investment&	content&	exposure&	mac\\
cost&	equity&	application&	fluctuation&	iphone ipad\\
income&	liability&	customer&	effect&	ipod touch\\
risk&	dividend&	intellectual property&	change&	device\\
hardware&	pricing&	service&	increase&	ipod\\
content&	transaction&	solution&	volatility&	os x\\
application&	payment&	operating system&	amount&	user\\ \hline
\end{tabular}
\label{tab:apple}
\end{table}

\subsection{Other implementation considerations}
\label{sec:implementation}
\paragraph{Normalization parameters, $\alpha$ and $\beta$}
\label{sec:normalization}
The word score computed in \eqref{eq:score} uses the normalization parameters $\alpha$ and $\beta$, which were fixed as $\alpha=\beta=1$ in the experiments. We briefly discuss how these may be modified to improve the performance of the algorithm in some cases. The score combines two different types of measures: depth and degree, which are distributed quite differently over the vocabulary. We need to ensure that the role both these measures play in identifying the core topics is ``balanced.'' In most documents, setting $\alpha=\beta=1$ works because degree is normalized by the logarithmic transformation. However, we find that a more robust score can be obtained by choosing $\alpha$ and $\beta$ so that after this power transformation the depth and degree measures both have median values equal to 1/2. Formally,  we choose $\alpha$ so that we have 
\begin{equation}
\text{Median}_{v \in V} \left(\frac{depth(v)}{\max_{u \in V} depth(u)}\right)^\alpha=\frac{1}{2}.
\end{equation}
The value for $\beta$ can be computed in an analogous fashion. For datasets in this paper, the values of $\alpha$ and $\beta$  so computed are quite close to one; incorporating this into the algorithm leaves all the results in the paper qualitatively unchanged. However, in our extensive experiments we did find some datasets for which using this normalization method improved performance. 

\paragraph{Performance}
In this paper, we have demonstrated that  \textit{Vec2Topic} works well both for user-generated and non-user generated datasets and even for single documents. The ability of the algorithm to capture  core topics depends on how well-defined these topics are. For documents with a clear core topic, the algorithm can extract it even when the size of the document is quite small. In our extensive experiments, we found it to perform well even for documents with a couple of hundred tokens in total, such as a news article. However, in that case, we did not build local word embeddings and instead used the knowledge-base word vectors alone. The reason this works is because a document such as a news article is written around a central theme or topic, and hence the context of usage is not that consequential for identifying the topic. 

We would like to point out that though  \textit{Vec2Topic} does not explicitly use a probabilistic approach, its output is not deterministic because it depends on building local word embeddings. In our experiments, we used the skip-gram approach, which results in slightly different word embeddings each time it is run. This effect is more pronounced with smaller  datasets, so while the degree score remains constant across runs, the depth score may vary slightly. Nevertheless, the overall results and the top core topics are qualitatively quite robust to this effect. 

The performance of the algorithm is also limited by the quality of the knowledge-base vectors. In our experiments, we used the English Wikipedia for our knowledge-base. One of the limitations therein was with respect to foreign words. This knowledge-base contains many foreign words (for instance, movie names) without sufficient context. This implies that the learned word representations of these words are not semantically accurate. To alleviate this issue, we removed such words (this issue came up with Mr. Kaminski's emails). Another related issue is that of people names. The word vectors for these tend to be quite close to each other, and if there are a large number of names, then this affects the depth measure. Clearly, people names would not make for key topics. So, to resolve this issue, in the email corpora, we remove names of people listed in the address fields. In the NIPS dataset, we remove the References section of the paper. Another way to resolve this issue is to run the dataset through a Named Entity Recognition algorithm.

\section{Conclusions}
In this paper we propose a novel technique for topic modeling that leverages understanding of word semantics using high-dimensional word vectors. We showed that  \textit{Vec2Topic} works well to extract a user's key topics of interest across their own generated content --- it also ranks these topics, and identifies keywords that best describe it. We contrasted it with the state-of-the-art topic modeling algorithm LDA, and observed that it works much better when the topic keywords are spread across the various documents, and are surrounded by several contextual words that are generic in nature.  Further we observe that the technique is not limited to user generated content; it works equally well on more structured documents such as  scientific papers, news articles, blogs, web pages, etc. It is also fairly robust to the corpus size --- it can scale from a single document to a large collection. 

One of our ongoing efforts is focused on extending the algorithm to identify phrases --- sequences of keywords that together capture the user's key interests. For example, turning to the example of a professional who works on datacenter servers, the phrase ``cloud server virtualization'' conveys a lot more topical context than cloud, server and virtualization individually. Each of these words may show up in several contexts in the user's data --- we therefore need a way to rank such phrases. Also, these phrases differ from bigram/trigram phrases in the sense that their words may not show up consecutively in the user's data. We believe extending topic modeling in this direction is an interesting avenue for future study.

\section{Acknowledgments}
The authors would like to thank Achal Bassamboo for many useful discussions.

%
\bibliographystyle{abbrv}
%
%

\end{document}